\documentclass[conference]{IEEEtran}
\IEEEoverridecommandlockouts
\usepackage{cite}
\usepackage{amsmath,amssymb,amsfonts}
\usepackage{algorithmic}
\usepackage{graphicx}
\usepackage{textcomp}
\usepackage{xcolor}
\usepackage{booktabs}
\def\BibTeX{{\rm B\kern-.05em{\sc i\kern-.025em b}\kern-.08em
    T\kern-.1667em\lower.7ex\hbox{E}\kern-.125emX}}

\begin{document}

\title{TeG: Temporal-Granularity Method for Anomaly Detection with Attention in Smart City Surveillance\\
\vspace{-0.3cm}
\thanks{This work was supported by the European ITEA SMART Mobility project on intelligent traffic flow systems.}
}

\author{\IEEEauthorblockN{Erkut Akdag, Egor Bondarev, Peter H.N. De~With}
\IEEEauthorblockA{\textit{VCA Group, Department of Electrical Engineering, Eindhoven University of Technology, The Netherlands}\\
P.O. Box 513, Eindhoven 5612AZ, The Netherlands \\ \{e.akdag, e.bondarev, p.h.n.de.with\}@tue.nl}
\vspace{-0.5cm}
}

\maketitle

\begin{abstract}
Anomaly detection in video surveillance has recently gained interest from the research community. Temporal duration of anomalies vary within video streams, leading to complications in learning the temporal dynamics of specific events. This paper presents a temporal-granularity method for an anomaly detection model~(TeG) in real-world surveillance, combining spatio-temporal features at different time-scales. The TeG model employs multi-head cross-attention~(MCA) blocks and multi-head self-attention~(MSA) blocks for this purpose. Additionally, we extend the UCF-Crime dataset with new anomaly types relevant to Smart City research project. The TeG model is deployed and validated in a city surveillance system, achieving successful real-time results in industrial settings.  
\end{abstract}


\section{Introduction}
\label{Intro}
Nowadays, closed-circuit television~(CCTV) systems for real-world surveillance provide many benefits to law enforcement agencies and municipal departments. One of the core objectives of real-world surveillance is abnormal behaviour detection to empower security systems, public safety, and operational efficiency. In the current industrial settings, the control-room operators are screening the increasing amount of CCTV footage, which is labor-intensive. Moreover, manual monitoring of surveillance data is prone to human errors, since operators should check multiple screens simultaneously. Therefore, automated systems are necessary to continuously analyze video feeds and detect unusual happenings in cities by reducing the human dependencies. 

Anomaly events are defined as deviations from the normal patterns. Although anomalies occur rarely, they have significant impact on public safety. Herewith, detection of anomalies in surveillance footage is one of the applications that have recently gained noticeable attention within the deep learning community. Our research is part of a Smart City project and aims to address a wide range of anomaly scenarios to enhance public safety. For instance, object throwing actions can indicate vandalism and potential violence, while traffic accidents require immediate attention to prevent further incidents and ensure timely medical assistance. Moreover, crime-based actions, such as fighting or gun fire, necessitate rapid intervention to protect citizens. Therefore, an anomaly detection surveillance system should address wide range of use cases to ensure comprehensive coverage of anomalies.

Another challenging issue overlooked in previous studies is the variable temporal duration of anomalies. In real-word surveillance, some anomaly incidents occur over prolonged periods, while others unfold in milliseconds. For instance, throwing actions and traffic accidents are short-duration anomalies, while the fighting and vandalism events last longer. The diversity between shorter and longer anomalies requires learning a model to consider a range of temporal scales. Neglecting the temporal dynamics does not allow a model to recognize the full spectrum of abnormal activities. Accordingly, a model should learn temporal granularity to efficiently cover abnormal events, regardless of their durations.

%update bold, and concat and res boxes.
\begin{figure*}[ht]
\centering
\includegraphics[width=0.99\linewidth]{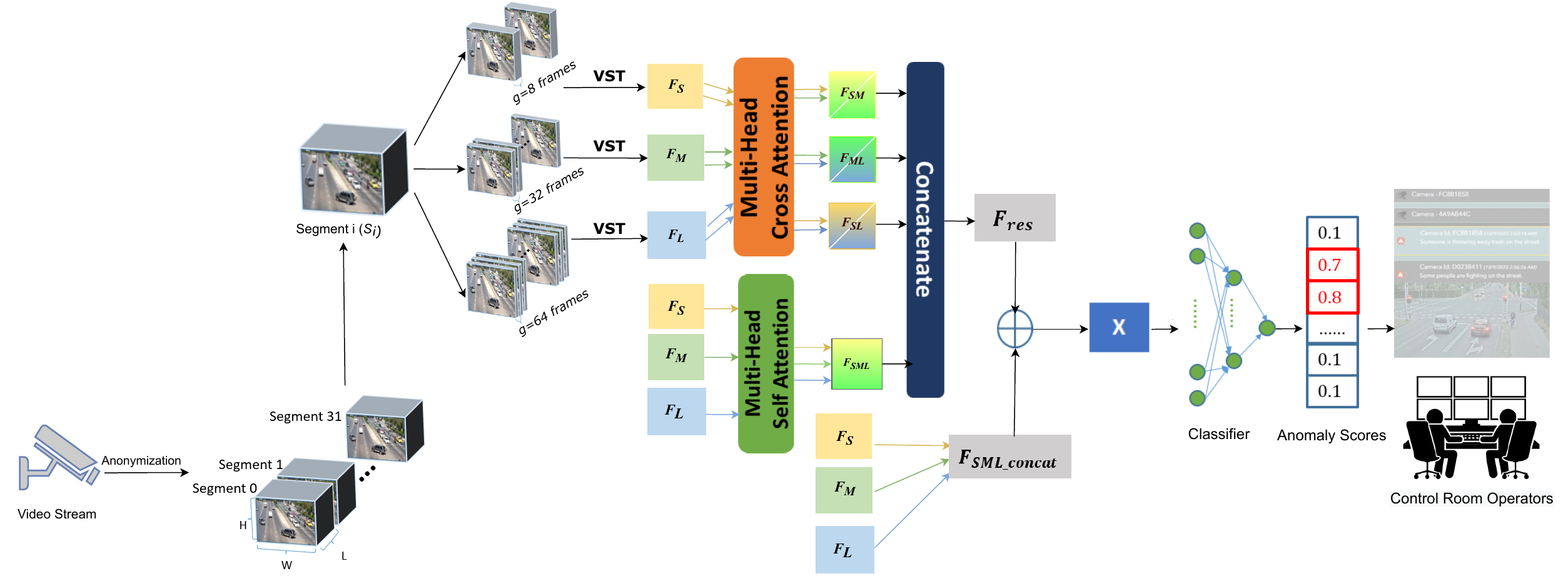}
\caption{Workflow of the TeG model. The input video is split into 32 video segments. The Video Swin Transformer~(VST) extracts features \textbf{$F_{S}$}, \textbf{$F_{M}$}, and \textbf{$F_{L}$} within short, medium, and long temporal granularity. Next, the TeG captures the correlations among three features and the dependencies among different video segments to fuse the features into the output feature matrix \textbf{$X$}. After anomaly scores of 32 video segments are obtained by a classifier, detailed information about the detected anomaly is provided to control-room operators.
\label{fig:proposed_method}}
\vspace*{-0.4cm}
\end{figure*}

To address the aforementioned challenges and hypothesis on temporal scaling, a temporal-granularity method~(TeG) is proposed to merge spatio-temporal features extracted by the Video Swin Transformer~(VST)~\cite{liu2022video} at different temporal scales. The TeG model learns the correlations between temporal granularity features using multi-head cross-attention~(MCA) and multi-head self-attention~(MSA) principles. Moreover, we extend the UCF-Crime~\cite{sultani2018real} dataset by including the anomaly types, defined in the previously mentioned Smart City project. Finally, the solution is deployed in a field lab, generating the detailed information about detected anomalies to control-room operators. The main contributions of this paper are as follows. 
\begin{itemize} 
\item Temporal-granularity method utilizing different spatio-temporal features extracted by the VST. 
\item Spatio-temporal feature fusion to learn correlations between different time-scale features within the MCA and MSA mechanisms with residual connection.
\item Deployment of the anomaly detection model in the city field lab, utilizing real-world surveillance data to obtain comprehensive coverage of various anomalies.
\end{itemize}

\section{Related Work}
There are~3 popular approaches for anomaly detection in literature. Unsupervised approaches learn the dominant data representation using regular surveillance data only. The studies~\cite{georgescu2021anomaly,liu2021hybrid, yuan2021transanomaly,feng2021convolutional,lee2022multi,zaheer2022generative,akcay2019ganomaly,yu2021abnormal,wang2010anomaly, wang2019gods, kim2021semi} reveal that these approaches are not efficient as borders between normal behaviour classes are undefined in the latent space. However, supervised approaches like~\cite{Petrocchi2021, Nasaruddin2020, Ullah2020} predict the presence of anomalies in the temporal domain only, while others~\cite{Zhou2019, Liu2019a} also locate the anomalies at the spatial level. This approach often outperforms unsupervised techniques for the specific anomalies for which they are trained. The important drawback is that these methods can only detect the anomalies existing in the dataset. Furthermore, annotating the training data is labor-intensive and expensive, as anomaly locations should be specified in every video frame or segment. Alternatively, weakly-supervised techniques based on multi-instance learning~(MIL)~\cite{tian2021weakly,sultani2018real,wu2021weakly,wu2021learning,deshpande2022anomaly,lv2021localizing,yan2022multiview, zhang2019temporal, wu2020not, li2022scale, zhang2022weakly, cao2022adaptive, sapkota2022bayesian, wu2022self, cho2023look, zhu2019motion, carbonneau2018multiple} have gained significant popularity because of reducing the annotation work by employing video labels. In this approach, train-set annotations only include an anomaly-class label. In contrast, test-set annotations in a video-class label require the starting and ending frame positions of an abnormal event in a video. This approach demands less labeled data than supervised learning, while providing higher accuracy than unsupervised methods.
% \begin{equation}
%     S_i^L=\sum_{j=0}^{(L/G)-1}c_j \ ,
% \end{equation}
\section{Proposed Method}
The TeG focuses on the temporal granularity concerning short, medium, and long durations of anomaly behaviour. Taking into account the variations in the attention focus of feature information across different video segment lengths, the TeG method aims to integrate essential features from multiple temporal scales, in order to improve the separation between the feature representations of abnormal and normal behavior segments. As shown in Fig.~\ref{fig:proposed_method}, each video clip is divided into 32 video segments to achieve a fair comparison with SotA models on public datasets. Each video segment, denoted as $S_i$, is later divided into "video chunks" based on the temporal-granularity variables. The parameter $S_i$ is defined by $S_i^L=\sum c_j $ where $j$ ranges within $0\leq j \leq (L/G)-1$, where $S_i^L$ is the $i^{th}$ video segment randomly picked up from 32~video segments with a video-segment length of $L$, $G$ is the temporal granularity expressed as the amount of levels representing temporal dynamics, and $c_j$ refers to a video chunk inside that segment. The temporal granularity parameter $G$ is empirically set to $8$, $32$, and $64$ frames for short, medium, and long temporal dynamics, respectively. The Video Swin Transformer~(VST)~\cite{liu2022video}, pre-trained on the Kinetics-400 dataset, extracts features at temporal granularity for all video chunks inside the video segment $S_i$. To form a single feature vector $F$ for a video segment, each feature matrix obtained by the video chunk is averaged along the temporal axis, according to:
\begin{equation}
    F=\bigcup_{i=0}^{31} \frac{\sum_{j=0}^{(L/G)-1} f_{\text{VST}} (c_j)}{L/G} \ ,
\end{equation}
%
% \begin{equation}
%     F=\bigcup_{i=0}^{31} \sum_{j=0}^{(L/G)-1} \frac{ f_{\text{VST}} (c_j)}{L/G} \ ,
% \end{equation}
%
where $F$ is the final output feature vector for an input video, concatenating 32 feature vectors obtained from segments, $f_{\text{VST}}(\cdot)$ is the processing function of the VST backbone. Also, for simplicity, feature vector $F$ is partitioned into \textbf{$F_{S}$}, \textbf{$F_{M}$}, and \textbf{$F_{L}$} producing short, medium, and long temporal-granularity feature representations, respectively. 

The TeG method employs the multi-head cross-attention~(MCA) principle~\cite{yan2022multiview} to $\mathbf{F_{S}}$, $\mathbf{F_{M}}$, and $\mathbf{F_{L}}$ feature representations with cross-correlating them and outputs the fused matrices $\mathbf{F_{SM}}$, $\mathbf{F_{ML}}$, and $\mathbf{F_{SL}}$. The multi-head self-attention~(MSA)~\cite{vaswani2017attention} principle uses $\mathbf{F_{S}}$, $\mathbf{F_{M}}$, and $\mathbf{F_{L}}$ to scale the features based on the global temporal correlations across video segments within the TeG method, resulting in a feature matrix $\mathbf{F_{SML}}$. 
All feature matrices $\mathbf{F_{SM}}$, $\mathbf{F_{ML}}$, $\mathbf{F_{SL}}$, and $\mathbf{F_{SML}}$ are concatenated, generating the $\mathbf{F_{res}}$, while in parallel, $\mathbf{F_{S}}$, $\mathbf{F_{M}}$, and $\mathbf{F_{L}}$ are concatenated to a feature matrix $\mathbf{F_{SML\_\textbf{concat}}}$. This processing is depicted in Fig.~\ref{fig:proposed_method}. The final fused-feature vector \textbf{X} is generated from the concatenated features with a residual connection.

The fused-feature matrix $\mathbf{X}$ is supplied to a 3-layer FCN classifier to output the anomaly probability scores for 32 video segments, as shown in Fig.~\ref{fig:proposed_method}. The loss function proposed by~\cite{tian2021weakly} is used for training the anomaly detection model (Equations~\eqref{eq1}-\eqref{eq3}), including loss functions for feature magnitude learning and classifier training.

The feature magnitude learning aims to increase the magnitude distance between the features in normal and anomalous videos. Driven by this, the top-$k$ MIL ranking loss is calculated, using two groups of top-$k$ features with the maximum magnitude in the feature matrices obtained from an abnormal and a normal input. As formulated in~\eqref{eq1}, the feature magnitude loss is defined by:
\begin{equation}
    \mathcal{L}_{\text{FM}} = \mathop{\max}(0, m-d_{\theta,k}(\mathbf{X}^{+}, \mathbf{X}^{-})) \ , \label{eq1}
\end{equation}
where $\mathcal{L}_{\text{FM}}$ is the feature magnitude learning loss function that maximizes the separability between the top-$k$ features from normal and abnormal videos, $m$ is a pre-defined margin, $\mathbf{X}^{+}$ denotes the fused abnormal video feature generated by the TeG~($\mathbf{X}^{-}$ for a normal video), as shown in Fig.~\ref{fig:proposed_method}, and $d_{\theta,k}$ calculates the difference between the mean of top-$k$ feature magnitudes in $\mathbf{X}^{+}$ and $\mathbf{X}^{-}$.

In the classifier training, the binary cross-entropy~(BCE) loss $\mathcal{L}_{\text{BCE}}$ is calculated with~\eqref{eq2}, given by: 
\begin{equation}
    \mathcal{L}_{\text{BCE}} = \sum_{s\in\mathbf{S}_{k}}-(y\log(\Bar{s}))+(1-y)\log(1-\Bar{s}) \ , \label{eq2}
\end{equation}
where $\mathbf{S}_{k}$ contains $k$ scores of the top-$k$ selected segments with maximum feature magnitudes, $\Bar{s}$ is the average of $s\in\mathbf{S}_{k}$, and~$y$ is the input label.

The total loss $\mathcal{L}_{\text{total}}$ for training the anomaly detection model is calculated by:
\begin{equation}
\begin{split}
    \mathcal{L}_{\text{total}} = & \mathcal{L}_{\text{BCE}} + \lambda_{\text{FM}}\mathcal{L}_{\text{FM}} \\
    &+ \lambda_{1}\sum\limits_{t=0}^{T-1}|s_{t}^{+}|^{2}+\lambda_{2}\sum\limits_{t=1}^{T-1}|s_{t}^{+}-s_{t-1}^{+}| \ ,  
    \label{eq3}
\end{split}
\end{equation}
where $\lambda_{\text{FM}}$, $\lambda_{1}$, and $\lambda_{2}$ are the weighting factors of each loss component, $s_{t-1}^{+}$ and $s_{t}^{+}$ represent the scores of two consecutive segments with timing $t-1$ and $t$ in an anomalous input, respectively. The summations $\sum\nolimits_{t=0}^{T-1}|s_{t}^{+}|$ and $\sum\nolimits_{t=1}^{T-1}|s_{t}^{+}-s_{t-1}^{+}|$ are added for sparsity and temporal smoothness constraints of the scores in anomalous videos, in addition to the $\mathcal{L}_{\text{FM}}$ in~\eqref{eq1}, and $\mathcal{L}_{\text{BCE}}$ in~\eqref{eq2}. 

\begin{table}[t]
\caption{Comparison of anomaly detection performances with existing methods on UCF-Crime, ShanghaiTech, and XD-Violence. 
\label{tab:SoTA}}
\begin{tabular}{l|c|c|c}
\toprule Method & UCF-Crime & ShanghaiTech & XD-Violence \\
 & AUC (\%) & AUC (\%) & AP (\%) \\
\midrule
Sultani et al.\cite{sultani2018real} &75.41& 86.30 & 73.20 \\
Wu et al.\cite{wu2020not}  & 82.44& - & 78.64\\
RTFM\cite{tian2021weakly} & 84.30& 97.21 & 77.81\\
Wu et al.\cite{wu2021learning} & 84.89& 97.48 & 75.90\\
Cao et al.\cite{cao2022adaptive}  & 84.67& 96.05 & - \\
MSL\cite{li2022self} & 85.30& 96.08 & 78.28 \\
Wu et al.\cite{wu2022self}& 85.99& 97.48 & 80.26 \\
Cho et al.\cite{cho2023look} & 86.07& 97.59 & 81.31 \\
BN-WVAD \cite{zhou2023batchnorm} & 87.24 & - & 84.93 \\
MGFN \cite{chen2023mgfn} & 86.67 & - & 80.11 \\
\midrule
TeG  &87.16& 95.32 & 84.57\\
\bottomrule
\end{tabular}
\vspace*{-0.4cm}
\end{table}

%Double check the formatting of the conference!!!

\section{Experiments and Field Validation}
\begin{figure*}[ht]
\centering
\includegraphics[width=0.95\textwidth]{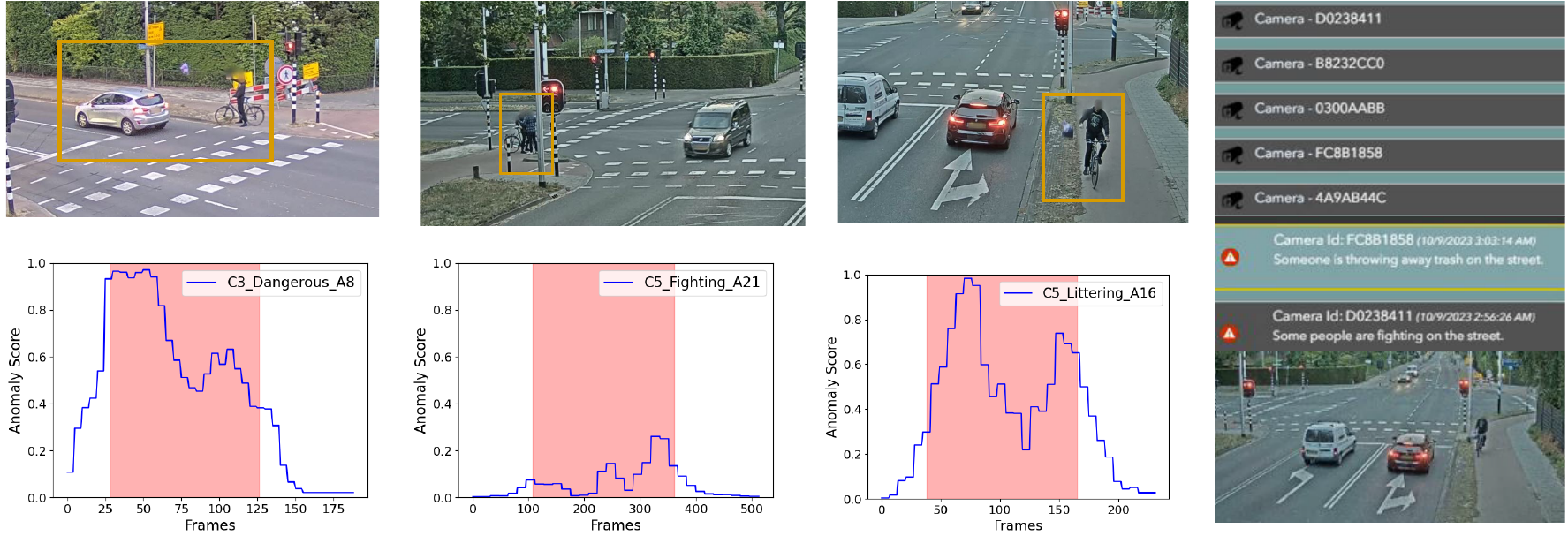}
\caption{Anomaly visualizations and the detection scores/frame of the TeG model. The three sub-figures at the left display anomalies: \textit{dangerous throwing}, \textit{fighting}, and \textit{littering}. The top figures illustrates the example frames sampled from the related anomalies, orange-colored boxes delineate where the anomaly occurs. The red-colored regions in the bottom graphs indicate the frame-level ground-truth labels of anomalous events. Right snapshot of the user interface indicating the cameras and informative text related to the detected anomaly. 
\vspace*{-0.4cm}
\label{fig:deployement}}
\end{figure*}

\subsection{Implementation Details}
The number of heads in the attention blocks is empirically set to 4. The anomaly detection model is trained by the Adam optimizer with a weight decay of 0.0005, a learning rate of 0.0001, and a batch size of~128 for 1,000~epochs. Each mini-batch consists of 64~randomly-selected normal and 64~abnormal videos. Everywhere, we set the margin $m$=100 and the number of snippets $k$=3 in the loss function by~\eqref{eq1}.

\subsection{SotA Results}
In Table~\ref{tab:SoTA}, the TeG model is compared with the SotA results on the UCF-Crime, ShanghaiTech, and XD-Violence datasets. The model achieves successful results on the UCF-Crime with 87.16\% area under the curve~(AUC), the ShanghaiTech with 95.32\% AUC, and the XD-Violence with 84.57\% average precision~(AP). It demonstrates better capability for identifying complex anomalies with various real-world scenarios, as presented in the UCF-Crime and XD-Violence datasets, compared to simpler anomalies in the ShanghaiTech dataset. The TeG outperforms other weakly supervised methods using the same loss function: RTFM~\cite{tian2021weakly} on the UCF-Crime and the XD-Violence datasets.

\begin{table}[b]
\vspace*{-0.3cm}
\caption{Amount of recorded anomalies. The anomalies are split into two categories: "unseen" and "seen", including three classes for each category.}
\centering
\begin{tabular}{l|c|c|c}
    \toprule
    \textbf{Anomaly category} & \textbf{Anomaly} & \textbf{Amount} & \textbf{Total}\\
    \midrule
    Unseen classes  
    ~ & Improper zone & 66\\
    ~ & Unlawful stop & 7 & \textbf{91}\\
    ~ & Improper turn & 18 \\ \midrule  
    Seen classes  
    ~ & Littering & 10\\
    ~ & Fighting & 2 & \textbf{18}\\
    ~ & Dangerous Throwing & 6 \\ \midrule  
    All-anomalies & - & - & \textbf{109} \\ \bottomrule
\end{tabular} 
\label{tab:anomalynumber}
\end{table} 

\begin{table}[t]
\vspace*{-0.1cm}
\caption{Performance of TeG on seen, unseen, and all-anomalies test sets. The normal and abnormal columns show the number of samples predicted as normal and abnormal classes. The accuracy results is reported along with $F_1$ scores.}
\centering
\begin{tabular}{l|c|c|c|c}
    \toprule
    \textbf{Category} & \textbf{Normal} & \textbf{Abnormal} & \textbf{Accuracy (\%)} & \textbf{$F_1$ score} \\
    \midrule
    Seen  & 3 & 15 & 83.33 & 0.90\\
    Unseen  & 19 & 72 & 79.10 & 0.88 \\
    All anomalies & 22 & 87 & 79.81 & 0.88\\    
    \bottomrule
\end{tabular} 
\label{tab:fp_anomaly}
\vspace*{-0.5cm}
\end{table} 

\subsection{Field-Lab Validation}
UCF-Crime~\cite{sultani2018real} is the largest dataset in literature with 13~classes included. However, the UCF-Crime is insufficient to cover real-world diversity of anomaly types. Therefore, we have extended the UCF-Crime dataset, including dangerous throwing~\cite{doi:10.2352/EI.2023.35.9.IPAS-286}, littering, and detailed traffic-accident classes, resulting in 17~classes in total. The TeG model is trained on this extended dataset and validated in the field lab to investigate real-world surveillance performance. For this, we have played anomalies~(``seen" and ``unseen") for 1 hour within the 5~cameras placed in the field lab, recording 25-fps video streams to test the TeG. The term ``seen" refers to the anomaly types on which the TeG has been trained, while ``unseen" refers to the classes on which the model has not been trained. The recorded ``seen" anomalies include~\textit{littering},~\textit{fighting}, and~\textit{dangerous throwing} actions. In addition, the recorded ``unseen" anomalies are \textit{improper zone}~(pedestrians and bicyclists on the road, vehicle parked on the curb), \textit{unlawful stop}~(car stopping at the center of the intersection), and \textit{improper turn}~(vehicles turning in the wrong direction, changing lanes, driving in a circle or doing U-turns). Table~\ref{tab:anomalynumber} depicts the number of ``seen" and ``unseen" anomaly types recorded from 5~cameras, regardless of whether one or more cameras capture the same abnormal behavior simultaneously. 

During validation, the first experiment is conducted on the seen classes to obtain anomaly scores per frame. The left part of Fig.~\ref{fig:deployement} demonstrates the output anomaly scores for selected anomaly videos. Notably, the TeG model showcases a high level of precision in distinguishing between normal and anomalous video segments. The anomaly scores, as depicted in the first and third bottom sub-figures, form a dominant peak, indicating the successful detections. Even in the other sub-figure, where the peak is less dominant, anomaly scores are still discernible, further reinforcing the model's efficacy in anomaly detection.

Second, we have approached the problem as binary anomaly detection to observe the performance of the TeG model on ``unseen" anomaly classes. All the abnormal behavior classes that the TeG predicts are considered as one ``anomaly" class, and the normal prediction is kept as a ``normal" class. As shown in Table~\ref{tab:fp_anomaly}, 3~out of~18 anomaly samples are classified as ``normal" instead of ``anomaly," giving an accuracy of 83.33\% and and 0.90~$F_1$ score, based on the experiments with the ``seen" test set. However, the TeG model achieves 79.10\%~accuracy and 0.88~$F_1$ score for the ``unseen" test set, showing a successful performance but relatively weaker results compared to ``seen" classes, as ``unseen" classes have not been trained during the training phase. In total, the TeG model incorrectly classifies 22~of 109~videos from the All-anomalies test set, resulting in 79.81\%~accuracy and 0.88~$F_1$ score. Therefore, in real-world surveillance, the proposed model can detect not only the ``seen" anomalies, but also the ``unseen" anomaly types, based on the results provided by Fig.~\ref{fig:deployement} and Table~\ref{tab:fp_anomaly}. Finally, the anomalies detected by the TeG method are sent individually through a secured HTTP endpoint to the control-room operators in real time. As shown at the right side of Fig.~\ref{fig:deployement}, the anomaly data packet contains the anomaly type, when it occurred, the camera it was detected by, and a GIF illustration of the anomalous action. 

\subsection{Execution Performance}
The execution performance of the TeG model is measured on a 25-fps video stream with a GTX-2080Ti GPU. For example, the feature extraction stage takes 104.96~ms for video segment of 2,133~ms length for each temporal granularity~(short, medium, long). This results in 104.96*3=314.88~ms for computing the final feature vector~\textbf{$X$}. The future fusion by attention mechanisms to the final output requires 2.5~ms. Therefore, in total, 317.38~ms are required for total computation time of that segment. This total divided by the segment duration is 14.87\% of the segment length in time, which also means approximately 1.5-sec. delay for a 10-sec. video segment. This is reasonable for a real-time surveillance application by recursively refreshing each segment's output.

\section{Conclusion}
This paper addresses the need for automated anomaly detection in CCTV surveillance to enhance public safety and operational efficiency. We have introduced a temporal-granularity method that merges spatio-temporal features across different temporal scales, thereby effectively capturing both short and prolonged anomalies. The proposed TeG model employs attention mechanisms to learn the correlations between features at various time-scales. We have also extended the UCF-Crime dataset to include additional anomaly types relevant to a European Smart City research project, providing a framework for further evaluation. The TeG validation in a field lab achieves real-time performance and demonstrates its practical applicability, providing detailed anomaly information to the control-room operators.

\end{document}